\documentclass{article}

\usepackage{arxiv}

\usepackage{amssymb}
\usepackage{setspace}
\usepackage{afterpage}
\usepackage{amsmath}
\usepackage{multicol}
\usepackage{lscape}
\usepackage{graphicx}%
\usepackage{multirow}%
\usepackage{longtable}
\usepackage{amsmath,amssymb,amsfonts}%
\usepackage{amsthm}%
\usepackage{mathrsfs}%
\usepackage[title]{appendix}%
\usepackage{xcolor}%
\usepackage{textcomp}%
\usepackage{manyfoot}%
\usepackage{booktabs}%
\usepackage{listings}%
\usepackage{comment}
\newcommand{\ra}[1]{\renewcommand{\arraystretch}{#1}}
\usepackage[linesnumbered,ruled,vlined]{algorithm2e}

\usepackage[utf8]{inputenc} 
\usepackage[T1]{fontenc}    
\usepackage{hyperref}       
\usepackage{url}            
\usepackage{booktabs}       
\usepackage{amsfonts}       
\usepackage{nicefrac}       
\usepackage{microtype}      
\usepackage{lipsum}         
\usepackage{graphicx}
\usepackage{natbib}
\usepackage{doi}
\usepackage{comment}

\usepackage{algorithmicx}%
\usepackage{algpseudocode}%

\title{GatedLexiconNet: A Comprehensive End-to-End Handwritten Paragraph Text Recognition System}

\newif\ifuniqueAffiliation
\uniqueAffiliationtrue

\ifuniqueAffiliation 
\author{ 		  
	\href{https://orcid.org/0000-0000-0000-0000}{\includegraphics[scale=0.06]{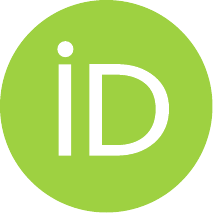}\hspace{1mm}Lalita Kumari}\\
	Department of Computer Science and Applications\\ 
	Panjab University, Chandigarh, India\\
	\texttt{lalita@pu.ac.in} \\
	\And
	\href{https://orcid.org/0000-0000-0000-0000}{\includegraphics[scale=0.06]{orcid.pdf}\hspace{1mm}Sukhdeep Singh} \\
	D.M. College (Affiliated to Panjab University, Chandigarh)\\
	Moga, Punjab, India\\
	\texttt{sukha13@ymail.com} \\
	\And
	\href{https://orcid.org/0000-0000-0000-0000}{\includegraphics[scale=0.06]{orcid.pdf}\hspace{1mm}Vaibhav Varish Singh Rathore} \\
	Computer Networking and Information Technology Division, PRL\\
	Ahmedabad, India\\
	\texttt{vaibhav@prl.res.in} \\
	\And
	\href{https://orcid.org/0000-0000-0000-0000}{\includegraphics[scale=0.06]{orcid.pdf}\hspace{1mm}Anuj Sharma}\\
	Department of Computer Science and Applications\\ 
	Panjab University, Chandigarh, India\\
	\texttt{anujs@pu.ac.in} \\
}
\else
\usepackage{authblk}

\newbox{\orcid}\sbox{\orcid}{\includegraphics[scale=0.06]{orcid.pdf}} 
\author[1]{%
	\href{https://orcid.org/0000-0000-0000-0000}{\usebox{\orcid}\hspace{1mm}Lalita Kumari\thanks{\texttt{lalita@pu.ac.in}}}%
}
\author[2]{%
	\href{https://orcid.org/0000-0000-0000-0000}{\usebox{\orcid}\hspace{1mm}Sukhdeep Singh\thanks{\texttt{sukha13@ymail.com}}}%
}
\author[3]{%
	\href{https://orcid.org/0000-0000-0000-0000}{\usebox{\orcid}\hspace{1mm}Vaibhav Varish Singh Rathore\thanks{\texttt{vaibhav@prl.res.in}}}%
}
\author[1]{%
	\href{https://orcid.org/0000-0000-0000-0000}{\usebox{\orcid}\hspace{1mm}Anuj Sharma\thanks{\texttt{anujs@pu.ac.in}}}%
}
\affil[1]{Department of Computer Science and Applications, Panjab University, Chandigarh, India}
\affil[2]{D.M. College (Affiliated to Panjab University, Chandigarh),
	Moga, Punjab, India}
\affil[3]{Computer Networking and Information Technology Division, PRL,
	Ahmedabad, India}
\fi

\renewcommand{\shorttitle}{\textit{arXiv} Template}

\hypersetup{
	pdftitle={A template for the arxiv style},
	pdfsubject={q-bio.NC, q-bio.QM},
	pdfauthor={David S.~Hippocampus, Elias D.~Striatum},
	pdfkeywords={First keyword, Second keyword, More},
}

\begin{document}
\maketitle

\begin{abstract}
The Handwritten Text Recognition problem has been a challenge for researchers for the last few decades, especially in the domain of computer vision, a subdomain of pattern recognition. Variability of texts amongst writers, cursiveness, and different font styles of handwritten texts with degradation of historical text images make it a challenging problem. Recognizing scanned document images in neural network-based systems typically involves a two-step approach: segmentation and recognition. However, this method has several drawbacks. These shortcomings encompass challenges in identifying text regions, analyzing layout diversity within pages, and establishing accurate ground truth segmentation. Consequently, these processes are prone to errors, leading to bottlenecks in achieving high recognition accuracies. Thus, in this study, we present an end-to-end paragraph recognition system that incorporates internal line segmentation and gated convolutional layers based encoder. The gating is a mechanism that controls the flow of information and allows to adaptively selection of the more relevant features in handwritten text recognition models. The attention module plays an important role in performing internal line segmentation, allowing the page to be processed line-by-line. During the decoding step, we have integrated a connectionist temporal classification-based word beam search decoder as a post-processing step. In this work, we have extended existing LexiconNet by carefully applying and utilizing gated convolutional layers in the existing
deep neural network. Our results at line and page levels also favour our new
GatedLexiconNet. This study reported character error rates of 2.27\% on IAM, 0.9\% on RIMES, and 2.13\% on READ-16, and word error rates of 5.73\% on IAM, 2.76\% on RIMES, and 6.52\% on READ-2016 datasets.
\end{abstract}


\section{Introduction}\label{sec1}

The recognition of handwritten text, known as the Handwritten Text Recognition (HTR) system, involves understanding handwritten content on digital surfaces or various media, including paper. This intricate problem has garnered significant attention from computer vision researchers in the community. HTR systems are categorized into two types: online and offline approaches \cite{srihari2000}. In online, the text is recognized on the go, which means we have geometric and temporal information about the written text. While in the offline mode, the text has already been written and further transcribed into the digital form. In this work, we have focused on offline HTR, and from now on the term HTR is referred to as offline HTR \cite{Tappert1990}. In the beginning, constrained by scarce resources, research focused on individual character recognition in text. However, with the advancement in computational capabilities over time, the HTR problem has seen a shift towards popularizing solutions that involve recognizing words and lines instead. In this current study, our novel approach involves the introduction of a closely integrated HTR system that is restricted by a lexicon and incorporates gated convolutional layers to regulate the flow of information. This design enables the system to proficiently recognize text from paragraphs in an end-to-end fashion without necessitating any external segmentation. The Hidden Markov Model (HMM) is mostly used in the initial phase of HTR studies \cite{yusof2003}. Due to the sequential nature of handwriting, variable length input sequences, sequential dependencies between symbols, and output as a sequence of symbols HTR is considered closely related to sequence learning. One of the drawbacks
of using HMM is their limited ability to capture long-range dependencies in sequences.
The HMMs are based on the Markov assumption, which states that the probability of
a particular state depends only on the preceding state. In the past few years, significant progress has been made in computational capabilities, leading to the emergence
of highly effective deep-learning models. Notably, the Convolutional Neural Network
(CNN), Recurrent Neural Network (RNN) and the fusion of both, known as Convolutional Recurrent Neural Network (CRNN), have demonstrated outstanding accuracy in HTR tasks \cite{shi2017}. This
work is a step forward towards the solution of the HTR. In the present work, a
segmentation-free approach of page-level HTR is discussed with gated convolutional
layers as a feature control mechanism. Therefore, through the development of the
system in this way, we can eliminate any latent inaccuracies and unnecessary computational expenses associated with segmentation and follow recognition methods. Gated convolutional-based NN has also given promising results in the past few years \cite{flor}. \par In this study, we introduce an effective end-to-end HTR system based on paragraph input. Our approach involves generating paragraph image features in a line-by-line fashion, employing a sophisticated combination of convolutional, gated convolutions, and depth-wise separable convolutional operations. These extracted features are subsequently fed into the Word Beam Search (WBS) decoder for further processing \cite{scheidl2024}. To ensure optimal performance, we adopted a two-step training strategy. Initially, we trained our NN model on the line level, focusing on a smaller and less complex task. Subsequently, we transferred the learned weights to the page-level model, enabling us to train the entire system end-to-end for enhanced efficiency. The underlying principle behind this methodology lies in leveraging the knowledge acquired from a simpler task to address more complex similar challenges. Following are the significant contributions made by the present study,

\begin{itemize}
    \item Our proposal involves a novel feature extractor that utilizes a combination of convolutional, gated convolutional, and depthwise separable convolutional layers. To the
best of our knowledge, this is the first time a gated convolutional layer has been
used to recognize paragraph text recognition.
\item This study leverages a novel deep learning architecture that incorporates gated convolutional layers within a Vertical Attention Network (VAN) \cite{Coquenet2022} for the training phase and integrates a Word Beam Search (WBS) \cite{scheidl2024} during decoding. The proposed GatedLexiconNet model demonstrates promising performance.
\item Our approach achieved impressive accuracy on state-of-the-art handwriting datasets. On IAM, the Character Error Rate (CER) is 2.27\% and Word Error Rate (WER) is 5.73\%. RIMES and READ-2016 followed suit, with CERs of 0.9\% and 2.13\% and WERs of 2.76\% and 6.52\%, respectively.
\item This study delves into the essential stages that constitute a generic HTR process to equip future readers with a firm foundation for grasping the intricacies of the proposed methodology.
\end{itemize} 

The subsequent sections of this paper are arranged as follows: Section 2, reviews the pertinent literature and discusses previous research efforts in the field. Section 3, provides an overview of our proposed system, outlining its key components and functionalities. Section 4,  details the experimental setup utilized to evaluate the performance of our system, including datasets, evaluation metrics, and implementation details. Section 5, delves into the variations observed among the proposed models and this study concludes in section 6, with a direction of future challenges in HTR.

\section{Related Work}\label{sec2}
In this section, we will examine the current state-of-the-art developments within the HTR domain. Our focus will primarily be on exploring techniques of offline HTR methodologies \cite{Seni1994}. Additionally, alternative methods have also been grounded in HMM. In a relevant research study, the system's input consists of individual words segmented into characters, enabling further processing utilizing the HMM \cite{kundu1988}. An alternative HMM-based approach explored the recognition of French cheques. This system employed meticulously designed features crafted from individual word attributes within the cheques \cite{KNERR1998}. Likewise, the fusion of HMM and RNN is utilized for word-level recognition tasks \cite{Senior1998}. The domain of HTR has also seen the application of Dynamic Programming (DP) techniques for word-level recognition. These approaches leverage path-finding algorithms to identify the most likely sequence of characters within a handwritten word. While DP-based methods have demonstrably improved character-level accuracy, the translation to overall word recognition accuracy is not always guaranteed \cite{chen99,bellman2015}. Additionally, the method of adaptive word recognition is implemented in tandem with manually crafted features \cite{Park2002}. One of the approaches involves the utilization of HMM-based statistical models, where each character is mapped to hidden states within the HMM \cite{Bunke2004}. A prominent strategy for word segmentation in handwritten text relies on the analysis of character spacing. This approach examines both the distances between characters within a word (intra-word) and the distances between characters belonging to different words (inter-word). By analyzing these spatial relationships, the system aims to identify word boundaries within the handwritten text accurately \cite{gatos2009}.An alternative method involves representing word images and text strings within a shared vector space, enabling the computation of distances between them. This transforms the task of recognition into one of retrieval \cite{almazan2014word}. \par  Early methods for addressing line-level recognition involve HMM. Techniques relying on statistical language models are employed for the recognition of entire lines \cite{Bunke2004}. In a related study, researchers combined HMM with stochastic context-free grammars to enhance the recognition of text lines \cite{Zimmermann2006}. Further, additional features are derived using a sliding window approach. Subsequently, the targeted text line is transformed into a series of timebased attributes using an HMM \cite{ploetz2009}. Promising outcomes are demonstrated by models that combine Hybrid HMM and Artificial Neural Network (ANN) approaches for recognizing text lines, along with the incorporation of slant and slope correction \cite{zamora2024}. HMMs have limitations in capturing intricate variations in handwriting styles, intricate connections between characters, and the context-dependent nature of handwriting. Additionally, HMMs might struggle with handling large vocabulary sizes and the inherent sequential nature of handwriting, as they rely on certain assumptions about the independence and Markovian properties of the data. In subsequent years, the employment of CNN-based systems has become a leading technique for extracting image features, while RNNs excel at retaining these features over extended periods. Consequently, the fusion of these methodologies has given rise to CRNN as a novel, cutting-edge approach. In a specific research endeavor, the focus was on line recognition, wherein CTC paired with BLSTM was utilized in conjunction with a token passing algorithm \cite{graves2009}. In one such study, the utilization of a CNN alongside MDLSTM and BLSTM is employed for the purpose of HTR \cite{Moysset2017,Wigington2017}. Introducing dropout enhances the accuracy of LSTM-based systems while retaining the advantageous sequential modelling capability of RNNs. Dropout, a regularization method, includes the random omission of certain hidden units in a neural network during training while maintaining their entirety during testing \cite{dropout2012,pham2014dropout}. The dropout offers a distinct advantage as it effectively addresses overfitting concerns. It serves as a regularizing method in nearly all intricate deep learning frameworks where the network is substantial. A particular research instance involves a multi-task network employed for concurrent script identification and text recognition. Within this study, an innovative type of LSTM unit known as Separable Multi-Dimensional Long Short-Term Memory (SepMDLSTM) is implemented to encode input text line images \cite{chen2017}. The CRNN structure amalgamates layers from Deep CNN and RNN, demonstrating exceptional outcomes in the identification of sequential items within images, particularly in the domain of scene text recognition \cite{shi2017}. which is expanded within the context of HTR \cite{SimpleHtr2018}.The CRNN model employed a substantial number of parameters for training, mainly due to the resource-intensive nature of RNNs. In a different research endeavour, an alternative system was introduced that relied on Gated CNN. This approach utilized fewer parameters and computational resources in comparison and managed to achieve cutting-edge outcomes \cite{flor}. In one of our earlier work we have extended this study by applying the depthwise seperable convolutional layer at suitable places and achieved competitive HTR results in text line recognition on standard datasets \cite{ivcnz}. The sequence-to-sequence recognition architecture also incorporates the language model, allowing adaptable information exchange between the external LM and the classifier. The external LM can also adjust to the training corpus, identifying commonly occurring errors during the training process. This methodology has been termed ’candidate fusion’ by the authors \cite{kang2019}. In one such study, a NN based language model has been incorporated into the decoding phase of HTR . This study investigated various HTR architectures alongside the NNLM during the decoding phase using sate-of-the-art datasets \cite{zamora2014lm}. \par In literature, limited efforts have been directed towards providing a comprehensive solution for text recognition at the paragraph level. Existing works have predominantly taken one of two distinct approaches. In first, segmentation techniques are employed to acquire paragraph-level images with distinct lines \cite{Vassilis2010}. In a particular segmentation approach, a method has been suggested for segmenting words that relies on the identification of the connecting region \cite{manmatha2024}.The same algorithm is also applied for the purpose of segmenting lines \cite{manmatha2005}. The statistical techniques for segmentation are also employed in tracing the cursive trajectory of an individual’s handwriting \cite{Arivazhagan2007}. In one of our earlier work we have used \cite{manmatha2024} word segmentation techniques at page level to identify each word one by one and then later used WBS \cite{scheidl2024} to improve the results further \cite{lalita2022}. The sequential process of segmentation and recognition is similarly utilized in a closely related field to handwritten text recognition, namely, scene-text recognition \cite{Weinman2014,Ye2015} and online HTR \cite{bengio95} in computer vision. After acquiring line images, optical recognition models operating at the line level are employed to extract transcriptions. However, these segmentation-based approaches introduced errors that led to subpar recognition outcomes. Alternatively, the second approach involves paragraph recognition without relying on external segmentation. In these paragraph recognition methods, either attention-based techniques are implemented for internal segmentation, or the task’s multi-dimensionality is addressed to achieve paragraph recognition in a unified process. In the initial strategy of segmentation-free methods, the complete document is transcribed in a continuous manner during the training process, incorporating an inherent line separation mechanism. The attention serves as the guiding element for vertically selecting between lines, while MDLSTMRNN is synchronized with CTC for accurate recognition of paragraph text \cite{bluche2017scan,bluche2017}. The CNN combined with BLSTM is employed for paragraph-level text recognition as an alternative to MDLSTM \cite{puigcerver2017}. The second method for recognizing paragraphs takes into account the two-dimensional nature of the task. It employs a Multi-Dimensional Connectionist Classification (MDCC) method to handle the two-dimensional features present in paragraph images. Through the utilization of a Conditional Random Field (CRF), a two-dimensional CRF is derived based on the accurate text information. This CRF structure serves as the controlling element for selecting lines from among numerous options \cite{Schall2018}. In a related study, a paragraph was treated as a single, extensive line of text. In this research, the approach involved integrating bi-linear interpolation layers with feature extraction layers from the FCN encoder. This fusion aimed to create an extended single line. The necessity for inserting line breaks during transcription and conducting line-level pre-training does not apply to this particular module \cite{YOUSEF2020-3}. In another study, A Vertical Attention Network (VAN) was used, integrating horizontal and vertical contextual information to enhance recognition accuracy. By combining convolutional and recurrent neural network layers, the model effectively captures intricate features in handwritten text. The vertical attention mechanism facilitates the selection of relevant information across lines, contributing to better understanding and context preservation \cite{Coquenet2022}. In their latest work, the authors presented an innovative end-to-end architecture, referred to as the Document Attention Network, designed for the task of recognizing handwritten documents without relying on segmentation. Alongside text recognition, the model can also assign labels to different sections of text, employing start and end tags in a manner reminiscent of XML. The architecture consists of an FCN encoder responsible for feature extraction and a series of transformer decoder layers that facilitate a step-by-step token prediction process \cite{dan2024}.
\section{System Overview}\label{sec3}
In this section, we provide an in-depth exploration of the components comprising our model architecture. Each element plays a crucial role in achieving our desired outcomes, from feature extraction to context understanding and decoding. We begin by outlining the core modules, including convolutional, gated convolutions and depthwise separable convolutional layers, which form the backbone of our model's feature extraction capabilities. Following this, we delve into the mechanisms of vertical attention and paragraph end detection, which are essential for contextual understanding and segmentation. Finally, we elucidate the workings of the WBS decoder, which converts learned features into final handwritten text. \par In our research study, we embarked upon a comprehensive exploration of IAM \cite{IAM}, READ\cite{read}, and RIMES \cite{rimes}  datasets, meticulously crafting our model's architecture to harness the rich information latent within paragraph images. As shown in figure \ref{fig:gcnn} Initially, our model operates at the paragraph level, employing preprocessing techniques to refine input images while leveraging pre-trained models trained on line-level data from the aforementioned datasets. Diverging from existing methodologies, our approach integrates gated convolutional layers into the initial phase, augmenting feature extraction capabilities. Subsequently, in the first part of the figure \ref{fig:gcnn} denoted as 1a, our model diverges from the VAN \cite{Coquenet2022} approach, deploying advanced techniques for enhanced feature extraction. Step 2 encompasses the utilization of depth-wise separable modules, complementing the preceding steps for comprehensive feature extraction. Notably, Step 3 introduces vertical attention mechanisms, facilitating precise feature localization at the line level within paragraph images, while concurrently detecting paragraph endings. Throughout the training, in step 4, LSTM networks serve as the decoder, maintaining consistency with established methodologies, with the implementation of the WBS decoder as step 5 during testing for optimal performance. It's imperative to note that while we introduced gated convolution layers, the underlying architecture and methodologies remained aligned with already established work VAN \cite{Coquenet2022}.
\subsection{Encoder Architecture}
As shown in Figure \ref{fig:gcnn} the encoder network leverages a combination of Convolutional Blocks (CB) and Depthwise Separable Convolution Blocks (DSCB) to capture a wider context for the final decision layer progressively. This configuration expands the receptive field to encompass an area of 961 pixels high and 337 pixels wide. Notably, both CB and DSCB blocks incorporate Diffused Mix Dropout (DMD) layers, a dropout strategy proposed in VAN \cite{Coquenet2022}. A CB consists of a sequential arrangement of two convolutional layers followed by gated convolutional layers followed by another convolutional layer and an Instance Normalization layer. Each convolutional layer and gated convolution layer utilizes 3x3 kernels with a ReLU activation function and zero padding to avoid information loss at the edges. While the first two convolutional layers maintain a stride of 1x1, the third layer employs varying strides for down-sampling. The use of a gated convolution layer further enhances the extracted features. This strategic use of strides reduces the feature map height by a factor of 32 and width by a factor of 8, leading to a significant decrease in memory requirements. Notably, DMD layers are integrated at three specific locations - following the activation functions of each convolutional layer within the CB.\par The DSCBs differ from CBs in two key aspects. Firstly, they employ Depthwise Separable Convolutions (DSC) [ in place of standard convolutional layers \cite{Chollet}. This substitution aims to achieve a reduction in the number of learnable parameters while maintaining comparable performance. Secondly, the final convolutional layer within a DSCB consistently utilizes a 1x1 stride. This approach ensures that the feature map dimensions are preserved throughout the encoder network until the final layer. Where feasible (i.e., when the feature map shapes remain consistent), residual connections with element-wise summation are implemented between consecutive blocks. These connections facilitate the efficient propagation of gradients during backpropagation, enhancing the parameter updates in the network's earlier layers as per the architecture of VAN \cite{Coquenet2022}.\par This section delves into the individual layers that make up Convolutional Blocks (CB) and Depthwise Separable Convolution Blocks (DSCB) within the encoder network (see Figure~\ref{fig:gcnn} for a visual reference). It provides a detailed examination of these components.

\subsubsection{Convolutional Layers}
Standard convolutional layers form the backbone of the encoder network. These layers are fundamental to CNNs, as they extract features by convolving a filter (kernel) of size 3x3 across the input data. To introduce non-linearity and enable the network to learn intricate relationships within the data, all convolutional layers in our CBs and DSCBs utilize ReLU activation functions. Furthermore, we employ zero-padding on the input data before the convolution process. This padding approach guarantees that the dimensions of the feature maps remain consistent throughout, preventing information loss at the borders. 
Equation \ref{eq:1}
shows the mathematical notation of a single convolution operation. We have three such convolution carefully placed inside a single CB block and this architecture as series of CBs similar to VAN \cite{Coquenet2022}.
\begin{equation}\label{eq:1}
    \text{out}(x, y) = \sum_{i=0}^{k} \sum_{j=0}^{k} \text{in}(x+i, y+j) \times \text{kernel}(i, j)
\end{equation}
\vspace{-1em}
where:
\begin{itemize}
\item [out(x, y)]: Output feature map at position (x, y).
\item [in(x, y)]:  Input feature map at position (x, y).
\item [kernel(i, j)]: Kernel weight at position (i, j).
\item [k]: Kernel size (assuming square kernel).
\end{itemize}

\begin{landscape}
\thispagestyle{empty} 
\begin{figure}[t]
    \centering
    \includegraphics[width=\linewidth]{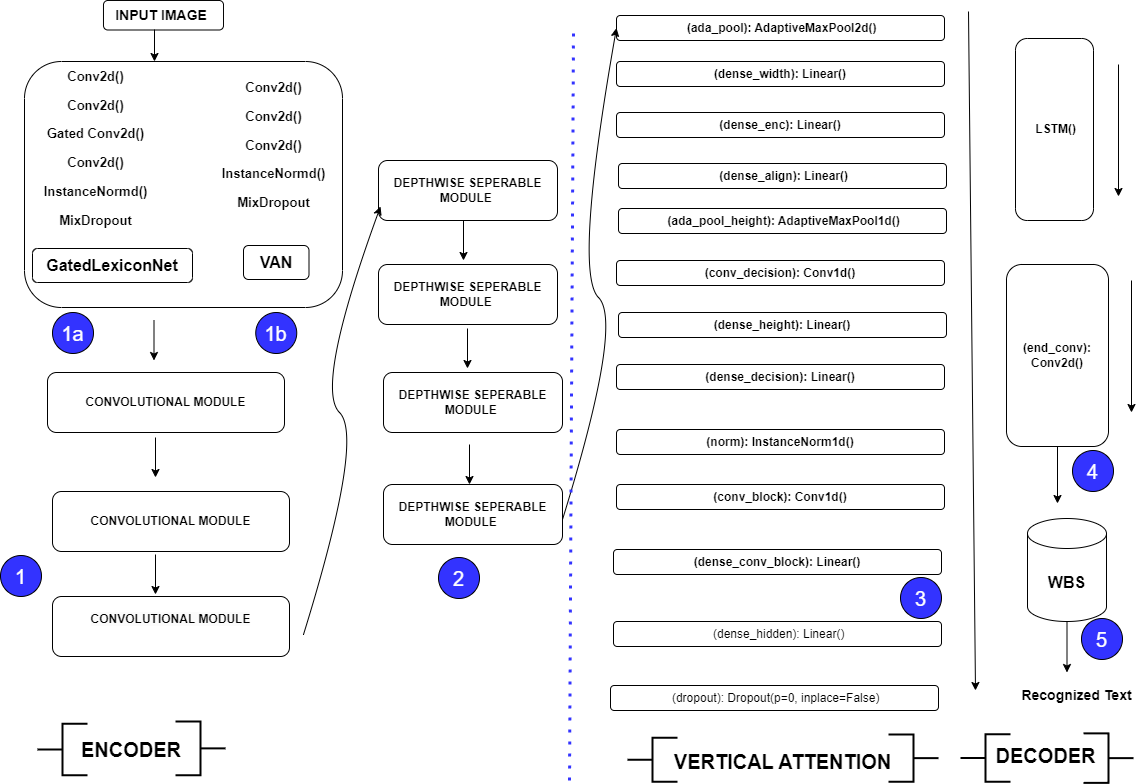} 
    \caption{System Overview of GatedLexionNet}
    \label{fig:gcnn}
\end{figure}
\end{landscape}

In this expression, $\text{out}(x, y)$ represents the value at position $(x, y)$ in the output feature map, which is obtained by convolving the input feature map $\text{in}$ with the kernel weights $\text{kernel}$. The summations iterate over all positions $(i, j)$ within the kernel size $k$, and the convolution operation computes the dot product between the input values and the corresponding kernel weights, followed by summing up the results.
\subsubsection{Gated Convolutional Layer}
A gated convolutional layer is a type of convolutional layer that incorporates gating mechanisms to control the flow of information through the layer. The gating mechanism is typically implemented using learnable parameters, allowing the network to adaptively adjust how much information is passed through.This work presents a novel contribution: gated convolutional layers integrated within the Convolutional Blocks (CBs). These layers extend the capabilities of standard convolutions by incorporating a gating mechanism. This mechanism empowers the network to discern informative features and selectively emphasize them, while simultaneously suppressing irrelevant information. Equation \ref{eq:gated_convolution} defines a gated convolutional layer, denoted as $\text{GatedConv2d}$, which consists of two convolutional operations: one for computing the filter values and another for computing the gating values. Here's the mathematical expression for the gated convolutional layer:

\begin{equation}\label{eq:gated_convolution}
\text{Output}(x, y) = \tanh(W_f \ast \text{Input}(x, y)) \odot \sigma(W_g \ast \text{Input}(x, y))
\end{equation}

where:
\begin{itemize}
\item $\text{Output}(x, y)$: Output feature map at position $(x, y)$
\item $\text{Input}(x, y)$: Input feature map at position $(x, y)$
\item $W_f$: Learnable filter weights
\item $W_g$: Learnable gate weights
\item $\ast$: Convolution operation
\item $\tanh$: Hyperbolic tangent activation function
\item $\sigma$: Sigmoid activation function
\item $\odot$: Element-wise multiplication
\end{itemize}

In Equation \ref{eq:gated_convolution}, the convolutional operations with $W_f$ and $W_g$ compute the filter values and gating values, respectively. The hyperbolic tangent activation function $\tanh$ squashes the filter values to the range $[-1, 1]$, while the sigmoid activation function $\sigma$ squashes the gating values to the range $[0, 1]$. The element-wise multiplication $\odot$ combines the filter and gating values, controlling the flow of information through the layer.

\subsubsection{Instance Normalization}
Instance Normalization (IN)   is a technique commonly used in deep learning, particularly in CNNs, to normalize the activations of intermediate layers within the network. The IN normalizes across the spatial dimensions of each individual sample independently. It aims to de-correlate the feature maps across the spatial dimensions, reducing the dependency between feature maps. This can lead to faster convergence during training and can improve the model's generalization ability.
 Following each convolutional, gated convolution layer within a CB, an Instance Normalization layer is applied carefully. Equation \ref{eq3} initializes an IN layer, denoted as $\text{norm\_layer}$, using the $\text{InstanceNorm2d}$ function. Here's its mathematical expression:

\begin{equation} \label{eq3}
\text{InstanceNorm2d}(X) = \frac{X - \mu}{\sqrt{\sigma^2 + \epsilon}} \times \gamma + \beta
\end{equation}

where:
\begin{itemize}
    \item $X$: Input feature map, where $C$ is the number of channels, $H$ is the height, and $W$ is the width.
    \item $\mu$: Mean of each channel computed across the spatial dimensions.
    \item $\sigma^2$: Variance of each channel computed across the spatial dimensions.
    \item $\epsilon$: Small constant added to the variance to prevent division by zero.
    \item $\gamma$: Learnable scaling parameter applied to each channel.
    \item $\beta$: Learnable shifting parameter applied to each channel.
\end{itemize}

The Instance Normalization operation performs channel-wise normalization on the input feature map. For each channel, it computes the mean $\mu$ and variance $\sigma^2$ across the spatial dimensions. Then, it normalizes the input using the mean and variance, scales it by $\gamma$, and shifts it by $\beta$. This normalization process helps stabilize the training process and improve the generalization ability of the model by reducing internal covariate shifts.

\subsubsection{Depthwise Separable Convolutional Layers}
For enhanced computational efficiency, DSCBs employ Depthwise Separable Convolutions (DSCs) in place of standard convolutional layers. DSCs achieve significant parameter reduction compared to standard convolutions by splitting the convolution process into two steps: a depthwise convolution applied independently to each input channel, followed by a pointwise convolution that reduces dimensionality across channels. This two-step approach leads to a more lightweight model, making it well-suited for resource-constrained environments.By decomposing the convolution operation into these two steps (depthwise and pointwise convolutions), the Depthwise Separable Convolutional layer achieves a significant reduction in the number of parameters and computation required, while still maintaining expressive power.   The Depthwise Separable Convolutional layer is composed of two stages: depthwise convolution (refer Equation \ref{eq4a}) and pointwise convolution (refer Equation \ref{eq4b}).

\begin{enumerate}
    \item \textbf{Depthwise Convolution:} 
    
    The first stage involves a depthwise convolution operation, denoted as $\text{depth\_conv}$, which convolves each input channel separately with its own set of filters. Here's its mathematical expression:
    
    \begin{equation} \label{eq4a}
    \text{depth\_conv}(X)_{i} = \sum_{c=1}^{C} X_{c} * K_{i,c}
    \end{equation}
    
    where:
    \begin{itemize}
        \item $X$ represents the input feature map of size $C \times H \times W$, where $C$ is the number of input channels, $H$ is the height, and $W$ is the width.
        \item $X_c$ represents the $c$-th input channel.
        \item $K_{i,c}$ represents the depthwise kernel corresponding to the $i$-th output channel and the $c$-th input channel.
        \item $\text{depth\_conv}(X)_i$ represents the $i$-th output feature map obtained by convolving the input feature map with the depthwise kernel $K_i$.
    \end{itemize}
    
    The depthwise convolution applies a single filter to each input channel independently. This operation helps reduce the computational cost by separating spatial and channel-wise filtering.
    
    \item \textbf{Pointwise Convolution:}
    
    The second stage involves a pointwise convolution operation, denoted as $\text{point\_conv}$, which performs a 1x1 convolution across all input channels to produce the final output. Here's its mathematical expression:
    
    \begin{equation} \label{eq4b}
    \text{point\_conv}(X) = \sum_{i=1}^{C} \text{depth\_conv}(X)_i * W_i
    \end{equation}
    
    where:
    \begin{itemize}
        \item $\text{depth\_conv}(X)_i$ represents the $i$-th output feature map obtained from the depthwise convolution.
        \item $W_i$ represents the learnable weights for the pointwise convolution.
        \item $\text{point\_conv}(X)$ represents the final output feature map obtained by convolving the depthwise output with the pointwise weights.
    \end{itemize}
    
    The pointwise convolution combines the information from different depthwise convolutions across channels to generate the final output. It allows for the creation of complex feature representations with fewer parameters compared to traditional convolutional layers.
\end{enumerate}

The Depthwise Separable Convolutional layer is commonly used in deep learning architectures to reduce the number of parameters and computational cost while maintaining expressive power.

\subsubsection{Residual Connections}
The network incorporates residual connections between consecutive blocks (CB or DSCB) whenever the feature map dimensions are preserved. This design choice, inspired by the VAN architecture \cite{Coquenet2022}, facilitates the efficient propagation of gradients during backpropagation. By directly adding the output of a previous layer to the input of the subsequent layer, these residual connections enable the network to learn increasingly intricate features.
\subsection{Attention}
Following the generation of a static two-dimensional feature representation by the convolutional layers, we introduce the vertical attention module. This module fulfils two critical functions:
\begin{enumerate}
    \item \textbf{Internal Line Segmentation}: It sequentially generates representations for individual text lines within the paragraph, adhering to the natural reading order (typically top-to-bottom).
    \item  \textbf{Paragraph End Detection} The module concurrently addresses the task of paragraph end detection, identifying the optimal stopping point for generating new line representations.
\end{enumerate}

 \subsubsection{Internal Line Segmentation}
This work leverages a soft attention mechanism, inspired by Bahdanau et al. \cite{bahdanau2024}, specifically a hybrid variant. Soft attention allows the model to focus on important features within the vertical sequence of features corresponding to each row in the image. The image is a grid, where each row holds features. The model calculates a score for each row, indicating how relevant it is for understanding a specific text line. These scores are then normalized to prioritize the currently processed row. Notably, the model focuses only on the vertical direction, allowing it to concentrate on specific rows crucial for the current text line. This approach reduces the complexity by assigning just one score per row. Ideally, one score should be close to 1, pinpointing the currently processed row, while others stay near 0. \par Through this iterative process for each line, the model builds line-level features. These features are essentially weighted combinations of the feature rows, with the weights determined by the previously calculated scores. This approach effectively captures the most distinguishing features of each text line. Importantly, the weighted sum ensures that the extracted line features remain a one-dimensional sequence. This characteristic is essential for using the well-established CTC loss function during training. CTC loss is particularly well-suited for tasks involving sequential data recognition, making it ideal for recognizing individual text lines in this scenario. \par The term "hybrid attention" signifies that the model calculates these scores by considering two sources of information. First, it captures the inherent meaning within the image features. Second, it encodes the spatial location of the features, enabling the model to prioritize relevant regions based on their position relative to the current text line.
\begin{equation} \label{eq5}
l_t=\sum_{i=1}^{H_f} \beta_{t, i} \cdot f_i
\end{equation}
As shown in Equation \ref{eq5} it involves the computation of attention weights ($\beta_{t, i}$) for each frame $i$ at attention step $t$. These attention weights sum to $1$ and quantify the importance of each frame for the attention step $t$. Each attention step $t$ is dedicated to processing the text line number $t$. The corresponding line features $l_t$ can be computed as a weighted sum between the feature rows and the attention weights. This weighted sum enables the line features $l_t$ to form a one-dimensional sequence, allowing the use of the standard CTC loss for each recognized text line.

\begin{equation} \label{eq7}
s_{t, i}=\tanh \left(W_f \cdot f_i^{\prime}+W_j \cdot j_{t, i}+W_h \cdot h_{W_f \cdot(t-1)}\right)
\end{equation}
The Equation \ref{eq7} computes the multiscale information $s_{t, i}$ for each row $i$ by combining various elements. $W_f \cdot f_i^{\prime}$ represents the contribution from the feature vector $f_i^{\prime}$ after transformation by weight matrix $W_f$. Similarly, $W_j \cdot j_{t, i}$ captures the influence of the context vector $j_{t, i}$ at attention step $t$, and $W_h \cdot h_{W_f \cdot(t-1)}$ represents the contribution from the decoder hidden state $h_{W_f \cdot(t-1)}$ after processing the previous text line. The $\tanh$ activation function ensures that the output is bounded within the range $[-1, 1]$. Overall, this equation integrates content-based and context-based information to compute the multiscale representation $s_{t, i}$ for each row $i$.

\subsubsection{Paragraph End Detection}
One of the main difficulties in paragraph recognition is figuring out how many lines of text there are without knowing beforehand. Since the process is sequential, the model needs a way to detect the end of a paragraph and stop creating new text lines. This work tackles this challenge by using a learning-based approach same as VAN \cite{Coquenet2022}. The goal is to teach the model when to stop recognizing new lines, essentially signaling the end of the paragraph. This is achieved by calculating a "paragraph end probability" at each step. By learning from these probabilities, the model becomes adept at identifying paragraph boundaries. However, this approach requires adding a new loss function specifically for end-of-paragraph detection alongside the standard CTC loss used for recognizing the text itself. During training, the final loss function combines both the CTC loss and this newly introduced cross-entropy loss.
\begin{equation}
\mathcal{L}_{\mathrm{ls}}=\sum_{k=1}^L \mathcal{L}_{\mathrm{CTC}}\left(p_k, y_k\right)+\lambda \sum_{k=1}^{L+1} \mathcal{L}_{\mathrm{CE}}\left(d_k, \delta_k\right)
\end{equation}
The CTC loss $\mathcal{L}_{\mathrm{CTC}}$ is computed between the predicted sequence $p_k$ and the target sequence $y_k$ for each frame $k$. This loss function measures the alignment between the predicted and target sequences, considering the variable-length nature of the sequences. Additionally, a cross-entropy loss $\mathcal{L}_{\mathrm{CE}}$ is applied to the decision probabilities $d_k$. This loss is summed over $L+1$ frames, where $L$ is the length of the target sequence. The decision probabilities represent the confidence scores for predicting line breaks. The ground truth decision probabilities $\delta_k$ are one-hot encoded, deduced from the line breaks in the target sequence. The value of  $\lambda$ is set to $1$.

\subsection{Decoder}
The decoder is essential for recognizing the sequence of characters from the extracted line features. Because the vertical attention mechanism has already converted it into a one-dimensional sequence, similar to a line image in standard HTR. \subsubsection{LSTM and Convolution-based Decoder}
 Following the vertical attention mechanism, a single layer with a predetermined number of cells (e.g., 256) called LSTM network is applied to the line features. The LSTMs are adept at handling sequential data like text, and in this case, the LSTM incorporates contextual information by processing the features along the horizontal axis (think of it as reading the line from left to right). For the first line, the LSTM starts with a clean slate (zero initial hidden states), but for subsequent lines, it carries forward information from the previous line to capture context across the entire paragraph. \par After the LSTM layer, a one-dimensional convolutional layer with a kernel size of 1 acts as a final transformation step. This layer takes the 256-channel output from the LSTM and converts it into a new representation with N+1 channels. N represents the number of characters the model can recognize, and the extra channel corresponds to a special symbol used by the CTC loss function (called the null symbol). Essentially, this final layer generates probabilities for each character (including the null symbol) at each position within the line.
 \subsubsection{WBS Decoder}
 To efficiently extract text from the model's output during testing, we leverage the WBS \cite{scheidl2024}. The model's output, initially a complex representation, is converted into a probability distribution for each character at each position in the line using a softmax function. This distribution essentially tells us how likely each character is at each spot.\par The WBS decoder utilizes a prefix tree built from the vocabulary of the test data. This tree essentially represents all valid word combinations in the expected language. By restricting its search to paths within this tree, the decoder significantly reduces the computational cost compared to methods that explore all possibilities. Ultimately, the WBS algorithm returns the most likely sequence of characters, forming the recognized text line (often referred to as the "beam"). In our experiments, we used the NGrams mode of the WBS algorithm with a beamwidth of 50 \cite{scheidl2024}.
 \section{Experimental Setup}
 \subsection{Datasets}
 To rigorously evaluate the performance of the proposed architecture, we employ three publicly available benchmark datasets: IAM \cite{IAM}, RIMES \cite{rimes}, and READ-2016 \cite{read}. A detailed description of each dataset is provided in the following subsections
 \subsubsection{IAM}
 The IAM handwriting dataset, released in 1999 and derived from the LOB corpus, is a valuable resource for training and evaluating text recognition systems. It offers a comprehensive collection of 1,539 scanned text pages containing 13,353 text lines and 115,320 labelled words. Notably, the dataset provides segmentation at various levels, including page, paragraph, line, and word. In our work, we specifically leveraged paragraph and line-level segmentation, following the train, test, and validation split detailed in Table \ref{tab:dataset}.
 \begin{table}[!hbt]
 \ra{1}
		\caption{Datsets splits in Training, Validation and Testing sets along with number of chracters each dataset containing}
		\begin{tabular}{lllll}
		Dataset	&      & Train & Validation & Test \\ \hline
			\multirow{2}{*}{\begin{tabular}[c]{@{}l@{}}IAM\\ (Number of Characters-79)\end{tabular}} & Line &6,482&976& 2,915\\  
			& Page & 747 &116&336      \\ \hline
			\multirow{2}{*}{\begin{tabular}[c]{@{}l@{}}RIMES\\ (\# of Characters-100)\end{tabular}} & Line & 10,532 &801 &  778 \\  
			& Page &  1400 & 100&100 \\ \hline
			\multirow{2}{*}{\begin{tabular}[c]{@{}l@{}}READ-2016\\ (\# of Characters-89)\end{tabular}}            & Line & 8,349 &1,040 & 1,138 \\  
			& Page &   1,584  & 179&  197    \\ \hline
		\end{tabular}
	
			\label{tab:dataset}
	\end{table}

\subsubsection{RIMES}
The RIMES dataset, standing for 'Recognition and Indexing of Handwritten Documents and Faxes' (Reconnaissance et Indexation de données Manuscrites et de fac-similés), provides a collection of handwritten letters sent by individuals to company administrators. This dataset comprises 12,723 pages containing 5,605 individual emails. The specific allocation of data for training, validation, and testing is outlined in Table \ref{tab:dataset}.

\subsubsection{READ-2016}
The READ-2016 dataset, introduced at the ICFHR 2016 conference \cite{read} as a component of the READ initiative, offers a valuable asset for historical handwritten document analysis. This dataset comprises a selection of documents from the Ratsprotokolle collection, encompassing approximately 30,000 pages of meeting minutes written in Early Modern German spanning the years 1470 to 1805. Notably, the dataset includes segmentation at the page, paragraph, and line levels, a feature provided by the READ-2016 project. The distribution of data for training, validation, and testing purposes is outlined in detail in Table~\ref{tab:dataset}.

\subsection{Preprocesisng}
\afterpage{%
  \vspace*{-2cm} 
}

\subsubsection{Preprocessing for Uniformity}
To ensure all datasets contribute data with consistent dimensions, a two-step preprocessing approach was applied to the images. First, images of varying sizes were scaled to a fixed size using bilinear interpolation, a method that maintains image quality during resizing.  Following this scaling step, zero-padding was employed to create a uniform image size of 800x480 pixels. This padding guarantees a minimum feature size of 100x15 pixels within the images. This minimum feature size is crucial for efficient feature extraction during the model's training phase.
\subsubsection{Data Augmentation to Mitigate Data Scarcity}
The process of gathering and labelling data can be quite complex. To tackle this challenge and effectively increase the available training data while introducing variability, we employed data augmentation techniques during the training phase. It's worth noting that these techniques were deliberately omitted during testing to ensure an impartial evaluation of the model's performance. The data augmentation procedure included introducing random modifications in various aspects of the images, such as resolution, perspective, elastic distortions (inspired by Youcef et al., 2024) \cite{yousef2024}, dilation and erosion, as well as brightness and contrast adjustments. Each transformation had a probability of 0.2 for application, and the sequence of transformations followed the specified order.

\subsection{Training Details}
\begin{singlespace}
\begin{algorithm}[!hbpt]

 \footnotesize
				\SetAlgoLined
				\KwIn{Batch of paragraph images \textit{Images}, and ground truth \textit{GroundTruth} with lines \textit{Line1}, \textit{Line2}, ..., \textit{LineL} and text corpus \textit{TextCorpus}}
				\KwResult{Training using backpropagation and Evaluation Metrics on given Dataset}
			
				\eIf{processing-mode="train"}{
					initializeFX()\; 
					initializeAttention()\;
					initializeLSTMDecoder()\;
					iteration=1\;
					features=computeFX(Images)\;
					$\alpha_{0}$=0, $h_{0}$=0, $\delta_{ctc}$=0, $\delta_{entropy}$=0\;
					\While{iteration$\le$L+1}{
						attention\_output, $\alpha_{\text{new}}$, decoder\_output=computeAttention(features, $\alpha_{\text{prev}}$, $h_{W_{f}(iteration-1)}$)\;
						\eIf{iteration$<$L+1}{
							prediction, Hidden\textsubscript{weights}=lstmDecode(attention\_output, Hidden\textsubscript{weights})\;
							$\delta_{ctc}$+=Loss\textsubscript{ctc}(prediction, GroundTruth)\;
							tempEndParagraph=$\begin{bmatrix}
							 0\\
							 1
							\end{bmatrix}$;
						}{
							tempEndParagraph=$\begin{bmatrix}
								1\\
								0
							\end{bmatrix}$;
						}
						$\delta_{entropy}$+=Loss\textsubscript{entropy}(decoder\_output, tempEndParagraph)\;
						iteration=iteration+1;\
					}
					backpropagation($\delta_{ctc}$, $\delta_{entropy}$)
					
				}{
					max\_line\_length=30, $\alpha_{0}$=0, $h_{0}$=0, decoder\_output=$\begin{bmatrix}
						0\\
						1
					\end{bmatrix}$, iteration=1, transcript=""\;
					initializeFX()\;
					initializeAttention()\;
					initializeLexicalDecoder()\;
					features=computeFX(Images)\;
					\While{iteration$\le$max\_line\_length \textbf{and} argmax(decoder\_output==1)}{
						attention\_output, $\alpha_{\text{new}}$, decoder\_output=computeAttention(features, $\alpha_{\text{prev}}$, $h_{W_{f(iteration-1)}}$)\;
						\If{argmax(decoder\_output==1)}{
							prediction, Hidden\textsubscript{weights}=lstmDecode(attention\_output, Hidden\textsubscript{weights})\;
							prediction=Softmax(prediction)\;
							Transcript=Concatenate(Transcript, lexicalDecode(prediction, TextCorpus))
						}
						iteration=iteration+1
					}
				}
				
				\caption{End-to-End Paragraph Recognition-Training and Testing}
				\label{algorithm:ch6balgo}	
			\end{algorithm}
   \end{singlespace}

The encoder and the final convolutional layer of the decoder in the VAN utilize pre-trained weights transferred from a previously trained architecture on official line-level segmented images within the same dataset. The training of this intermediate architecture is focused solely on the recognition component of the task. This two-stage approach is designed to expedite the convergence of the VAN and improve its recognition performance.

The pre-training process employs the same data augmentation strategies and pre-processing steps as the subsequent VAN training. The VAN takes whole paragraph images as input. The ground truth is segmented into a sequence of text line transcriptions using line break annotations, enabling the application of the CTC loss function during VAN training. This loss function facilitates line-by-line alignment between the recognized text lines and the ground truth. The following algorithm details the operation of the proposed model in a line-by-line manner.

\begin{longtable}{p{3cm}p{10cm}}
 \ra{1.5}
\textbf{Line 1:} & Verification of the processing mode to determine the operation of the decoder.\\
\textbf{Lines 2-4:} & Initialization of feature extraction (initializeFX()), attention mechanism (initializeAttention()), and LSTM decoder modules (initializeLSTMDecoder()) respectively.\\
\textbf{Line 6:} & Compute two-dimensional features for a batch of training images (computeFeatures()).\\
\textbf{Line 7:} & Initialization of attention weights (\textit{$\alpha_0$}), decoder's hidden state (\textit{$h_0$}), cross-entropy loss (\textit{$\delta_{entropy}$}), and CTC loss (\textit{$\delta_{ctc}$}).\\
\textbf{Lines 8-9:} & Calculation of one-dimensional line features (\textit{$l_{t}$}), computation of attention weights (\textit{$\alpha_{t}$}) using the attention mechanism (computeAttention()), and determination of the probability of detecting the end of a paragraph (\textit{$d_{t}$}).\\
 \textbf{Line 11:} & Calculate the probability of the predicted sequence (\textit{$prediction_{t}$}) for each line feature (\textit{$l_{t}$}).\\
\textbf{Lines 12-17:} & Calculate \textit{$\delta_{ctc}$} and \textit{$\delta_{entropy}$} by comparing \textit{$d_{t}$} with the one-hot encoding (\textit{$tempEndParagraph_{t}$}) representing the known ground truth of paragraph ends.\\ 
\textbf{Line 20:} & Perform backpropagation of losses to update the model weights.\\
\textbf{Line 27:} & If in testing mode, initialize the maximum number of lines (\textit{$max_line_length$}) that a paragraph in the dataset contains, along with \textit{$\alpha_{0}$}, \textit{$h_{0}$}, \textit{$d_{t}$}, and the page transcription as blank.\\
\textbf{Lines 23-25:} & Initialization of the feature extraction, attention mechanism, and LSTM decoder modules respectively.\\
\textbf{Line 26:} & Extract two-dimensional features for a batch of test images \textit{$Images$}.\\
\textbf{Lines 27-30:} & Iterate until the total number of processed lines is less than \textit{$max_line_length$} or the value of \textit{$d_{t}$} does not indicate the end of a paragraph. During each iteration, the attention module recurrently generates the features \textit{$l_{t}$}, and the LSTM decoder outputs the probability of characters.\\
\textbf{Line 31:} & Apply softmax function to the output of the LSTM decoder to transform it into the format required by the lexical decoder (Softmax()).\\
\textbf{Lines 32-36:} & For each feature \textit{$l_{t}$}, the lexical decoder algorithm computes predictions using the text corpus and concatenates the results in a line-by-line manner.\\
\end{longtable}

\subsection{Evaluation Metric}
This investigation utilizes cutting-edge metrics, namely CER and WER, to evaluate the efficacy of the proposed model. Even minor inaccuracies in transcription can significantly affect the overall statistical analysis. For instance, a simple substitution like mistaking "u" for "v" underscores the critical need for meticulous evaluation, given that research systems ultimately serve real-world applications.

CER is computed utilizing the Levenshtein Distance (represented by $L_d$) between the ground truth text sequence ($g$) and the predicted text sequence ($p$). To account for the varying impact of errors within longer lines on the overall error rate, we normalize CER by the total length of the line. Similarly, WER evaluates errors at the word level rather than the character level, as elaborated in Equation (\ref{eq:new_cer})

\begin{equation}
\text{CER} = \frac{\sum_{j=1}^{N} L_d(g_j, p_j)}{\sum_{j=1}^{N} |g_{j,\text{len}}|} \label{eq:new_cer}
\end{equation}

\subsection{Lexicon based WBS Decoding}
This section explores the application of the lexicon-driven WBS decoding algorithm (Scheidl et al., 2024) within the context of our model. During the decoding stage, the algorithm initializes with an empty candidate set, which represents potential text sequences at a specific time step. Concurrently, a prefix tree is constructed from the available text corpus. In our case, the corpus comprised text extracted from images belonging to the dataset split that is currently being processed (either test or validation). At each time step $i$, the prefix tree is queried to identify all valid word options present in the corpus. Subsequently, the candidate set is expanded by incorporating these potential word continuations. This iterative process continues from $i=0$ to $i=I$, where $I$ represents the total number of time steps. The state of the candidate set, denoted as either "word" or "non-word," determines the allowable character additions during expansion. Transitioning to a "non-word" state occurs when characters not found in the lexicon are included in the candidate set. Only candidate sets in the "word" state undergo expansion with characters suggested by the prefix tree, ensuring that these characters form valid word prefixes. In contrast, candidate sets in the "non-word" state can be extended with both word and non-word characters. This study employed the "NGrams" operation mode of the decoding algorithm with a beam width of 50.
\subsection{Results}
The soft attention mechanism used in this study,  successfully captured line-level features, enabling the model to achieve high accuracy in recognizing individual text lines within an image. The incorporation of gated convolutional layers demonstrably improved the effectiveness of the model in extracting discriminative features from the input image. This enhancement in feature learning translated to a significant improvement in text recognition accuracy, particularly for complex or challenging paragraph layouts. The WBS decoder further significantly improved the efficiency of the text recognition process by leveraging a prefix tree and reducing the search space. The combination of gated convolutional layers, the attention mechanism and the WBS decoder yielded a robust and efficient text recognition system, demonstrating its effectiveness in handling paragraph recognition tasks.\par Compared to baseline models of VAN, as shown in Table \ref{table:resl} our proposed architecture achieved a percentage improvement of 54.69\%  in CER and 65.25\% in WER of IAM dataset, 70.39\% in CER and 66.82\% in WER of RIMES dataset and 40.66\% in CER and 53.22\% in WER of READ-2016 dataset. This improvement suggests that the gating mechanism effectively selected the most relevant features for each text line, leading to more robust recognition performance. Our results suggest that gated convolutional layers play a crucial role in learning informative line-level features. This finding underscores the importance of incorporating such mechanisms for accurate paragraph recognition, especially in scenarios with varying text complexities.
\begin{table}[!htbp]
					\ra{1.5}
					\caption{Results and Comparison Table}
					\begin{tabular}{cllll}
						\hline
						\textbf{Reference}       & \multicolumn{1}{c}{\textbf{\begin{tabular}[c]{@{}c@{}}CER(\%)\\ Valid\end{tabular}}} & \multicolumn{1}{c}{\textbf{\begin{tabular}[c]{@{}c@{}}WER(\%)\\ Valid\end{tabular}}} & \multicolumn{1}{c}{\textbf{\begin{tabular}[c]{@{}c@{}}CER(\%)\\ Test\end{tabular}}} & \multicolumn{1}{c}{\textbf{\begin{tabular}[c]{@{}c@{}}WER(\%)\\ Test\end{tabular}}} \\ \hline
						\multicolumn{5}{c}{\textbf{IAM Dataset}}                                                                                                                                                                                                                                                                                                                                           \\ \hline
						\multicolumn{1}{l}{CNN+MDLSTM\cite{bluche2017scan}} & --                                                                                 & --                                                                                    & 16.2                                                                                & --                                                                               \\ \hline
						\multicolumn{1}{l}{CNN+MDLSTM\cite{bluche2016}} & 4.9                                                                                & 17.1                                                                                   & 7.9                                                                                & 24.6                                                                              \\ \hline
						\multicolumn{1}{l}{RPN+CNN+BLSTM+LM\cite{curtis2018}} & --                                                                               & --                                                                                   & 6.4                                                                                & 23.2                                                                              \\ \hline

						\multicolumn{1}{l}{VAN\cite{Coquenet2022}} & 3.02                                                                                  & 10.34                                                                                    & 4.45                                                                                & 14.55                                                                               \\ \hline
						
											\multicolumn{1}{l}{LexiconNet \cite{lalita2023}} & 1.89                                                                                    & 5.17                                                                                    & 3.24                                                                               & 8.29                                                                             \\ \hline
           \multicolumn{1}{l}{GatedLexiconNet (Present Work)} & 1.43                                                                                    & 3.74                                                                                    & \textbf{2.27}                                                                                 & \textbf{5.73}                                                                             \\ \hline
						
						\multicolumn{5}{c}{\textbf{RIMES Dataset}}                                                                                                                                                                                                                                                                                                                                         \\ \hline
						\multicolumn{1}{l}{CNN+MDLSTM\cite{bluche2016}}     &  2.5                                                                                    &  12.0                                                                                    &   2.9                                                                                  &12.6                                                                                     \\ \hline
						\multicolumn{1}{l}{RPN+CNN+BLSTM+LM\cite{curtis2018}} & --                                                                               & --                                                                                   & 2.1                                                                                & 9.3                                                                              \\ \hline
						\multicolumn{1}{l}{VAN\cite{Coquenet2022}} & 1.83                                                                                  & 6.26                                                                                    & 1.91                                                                                & 6.72                                                                               \\ \hline
						
						\multicolumn{1}{l}{LexiconNet\cite{lalita2023}} & 1.06                                                                                   & 2.91                                                                                    & 1.13                                                                                 & 2.94                                                                             \\ \hline
      \multicolumn{1}{l}{GatedLexiconNet (Present Work)} & 0.88                                                                                    & 2.30                                                                                   & \textbf{0.9}                                                                                 & \textbf{2.76}                                                                             \\ \hline

						\multicolumn{5}{c}{\textbf{READ-2016 Dataset}}                                                                                                                                                                                                                                                                                                                                     \\ \hline
						\multicolumn{1}{l}{VAN\cite{Coquenet2022}} & 3.71                                                                                  & 15.47                                                                                    & 3.59                                                                                & 13.94                                                                               \\ \hline

						\multicolumn{1}{l}{LexiconNet\cite{lalita2023}} & 2.29                                                                                 & 7.46                                                                                    & 2.43                                                                                & 7.35                                                                          \\ \hline
      \multicolumn{1}{l}{GatedLexiconNet (Present Work)} & 2.08                                                                                    & 6.13                                                                                   & \textbf{2.13}                                                                                 & \textbf{6.52}                                                                             \\ \hline

					\end{tabular}
		\label{table:resl}	
			\end{table}

\clearpage

\section{Discussion}
Our study investigated the impact of gated convolutional layers on paragraph recognition accuracy. We conducted comparisons among three approaches, as illustrated in Figure \ref{figure:ch7fig-2}:
\begin{figure}[!hbpt]
\centering
		\includegraphics [width=\textwidth, height=12cm]{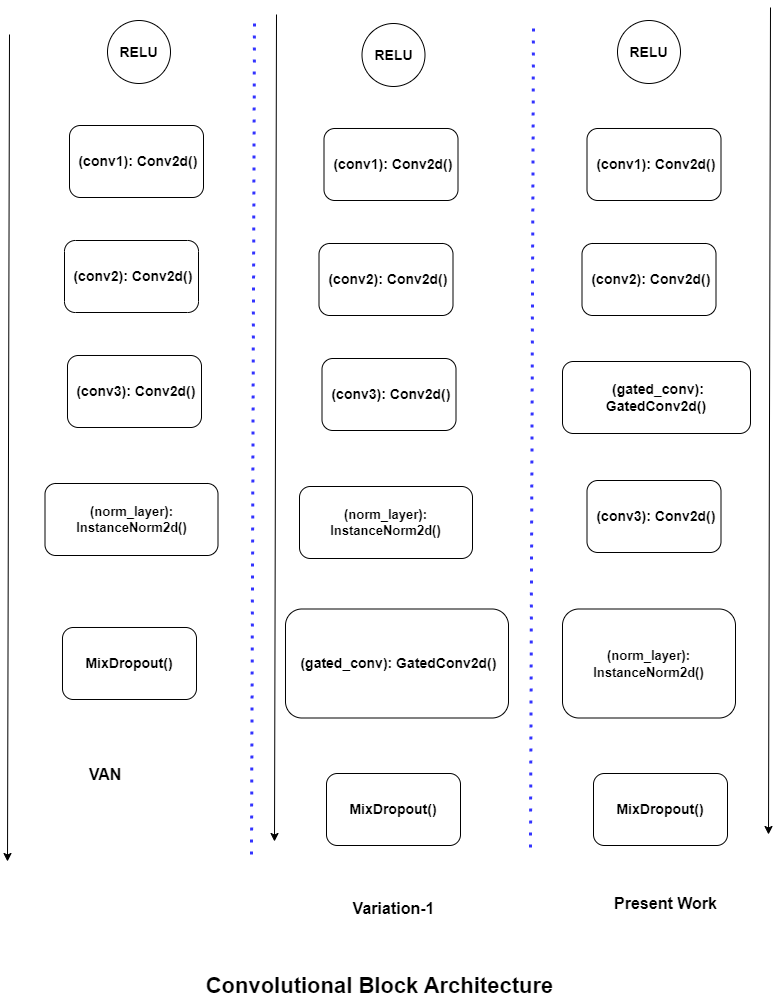}
		\caption {Convolutional block architecture of present work compared with VAN}
		\label{figure:ch7fig-2}
\end{figure}

\begin{itemize}
    \item \textbf{VAN Baseline}: This approach represents the vanilla VAN architecture comprising three convolutional layers (conv1, conv2, conv3), followed by normalization and dropout layers.
    \item \textbf{Variation 1: Late Gated Layer Integration}: In this variation, we introduced a gated convolutional layer after conv3 and the normalization layer of VAN. However, this approach encountered "NaN" values during training, impeding convergence.
    \par We postulate that incorporating the GCN at this late stage may have exacerbated the vanishing/exploding gradient problem, particularly with high dimensionality. The intricate interactions between the late GCN and the subsequent layers might have led to unstable gradients, resulting in the model outputting NaN values.
    \begin{figure}[!hbpt]
	\includegraphics [width=\textwidth, height=6cm]{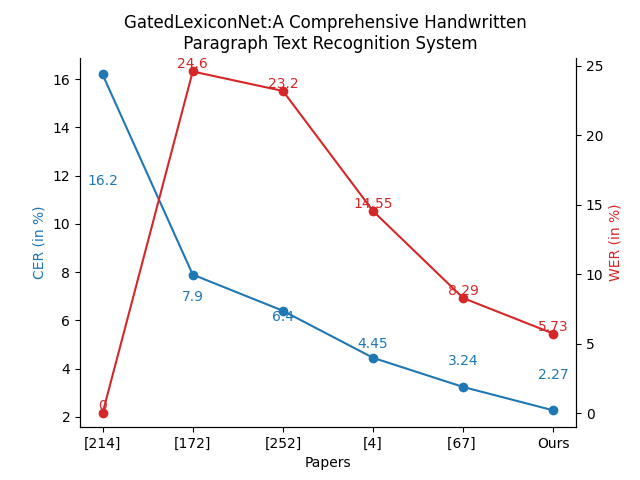}
	\caption {Comparison graph of present work with other state-of-the-art line level works on IAM Dataset}	
	\label{figure:cmpch71}
\end{figure}
    \item \textbf{Proposed Model: Early Gated Layer Integration}: To tackle this issue, our proposed model integrates a gated layer after conv2 and the normalization layer. This strategic placement notably enhanced the model's performance without encountering NaN values during training. We attribute this improvement to introducing the gated layer earlier in the network, allowing it to learn more informative features at a lower dimensionality, thereby mitigating the vanishing/exploding gradient issue. Consequently, this led to a more stable training process and improved recognition accuracy.
\end{itemize}

\begin{figure}[!hbpt]
	\includegraphics [width=\textwidth, height=5cm]{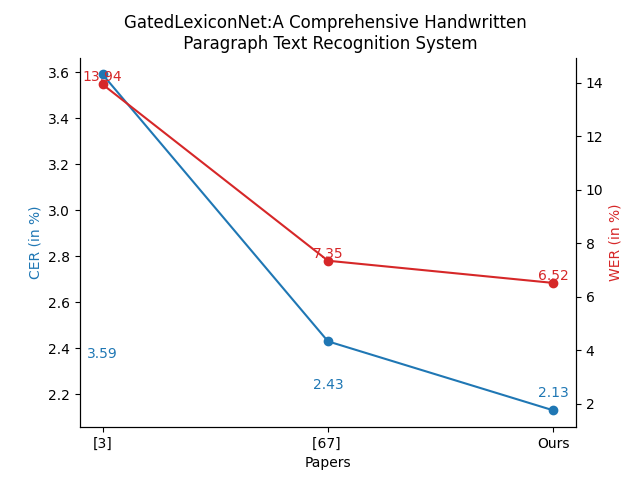}
	\caption {Comparison graph of present work with other state-of-the-art line level works on READ-2016 Dataset}	
	\label{figure:cmpch73}
\end{figure}

\begin{figure}[!hbpt]
	\includegraphics [width=\textwidth, height=6cm]{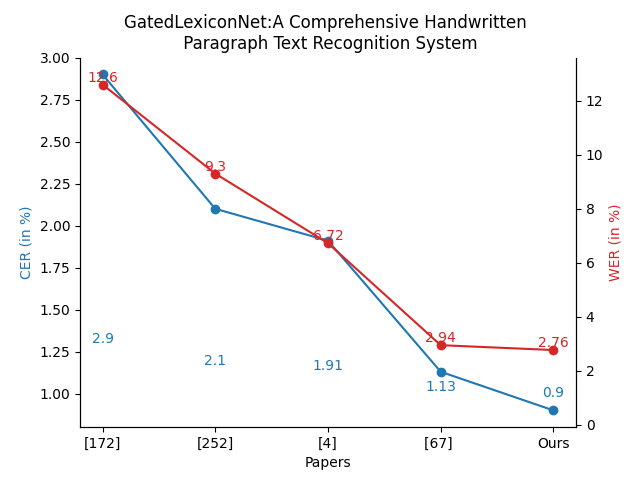}
	\caption {Comparison graph of present work with other state-of-the-art line level works on RIMES Dataset}	
	\label{figure:cmpch72}
\end{figure}
These findings underscore the significance of strategic gated layer placement within the network architecture. Our results indicate that incorporating gated layers earlier in the convolutional block can effectively address the curse of dimensionality and enhance model performance. Evaluation on IAM, RIMES, and READ-2016 datasets (refer to Figures \ref{figure:cmpch71}, \ref{figure:cmpch72}, and \ref{figure:cmpch73}) reveals substantial enhancements in CER and WER compared to the results presented in Table \ref{table:resl}. It is noteworthy that a '0' value in the graphs indicates that the metric was not reported in the corresponding study.

\section{Conclusion}
This work introduces GatedLexiconNet, a novel end-to-end paragraph recognition system that leverages a lexicon-based approach. The system eliminates the need for explicit text line segmentation and incorporates state-of-the-art techniques from HTR. Notably, a WBS lexicon decoder is integrated within the base model as a post-processing step. The core contribution of this work lies in modifying the encoder's convolutional block by carefully embedding a gated convolutional layer. This architectural change demonstrably improves both CER and WER on established benchmark datasets (IAM, RIMES, READ-2016). Additionally, a comprehensive analysis of the design choices and their impact on performance is provided.
GatedLexiconNet exhibits proficiency in recognizing complex layouts containing slanted text lines. It is important to note that the current system is designed specifically for single-column handwritten text layouts. Achieving end-to-end page recognition remains a computationally expensive challenge. Future work will focus on mitigating the computational complexity associated with this task.

\section*{Acknowledgement}
This research is funded by the Government of India, University Grant Commission, under the Senior Research Fellowship scheme. We are thankful to the authors of \cite{Coquenet2022} and \cite{scheidl2024} for their work to follow at GitHub. The computations were performed on the Param Vikram-1000 High Performance Computing Cluster of the Physical Research Laboratory (PRL)(URL-https://www.prl.res.in/prl-eng/paramvikram1000)

\bibliographystyle{unsrtnat}
\bibliography{references}

\end{document}